%% file: main.tex
\documentclass[11pt]{article}

\usepackage[final]{acl}

\usepackage{times}
\usepackage{latexsym}

\usepackage[T1]{fontenc}

\usepackage[utf8]{inputenc}
\usepackage{amsmath}
\usepackage{amssymb}
\usepackage{amsfonts}

\usepackage{latexsym}
\usepackage{inconsolata}

\usepackage{booktabs}
\usepackage{tabularx}
\usepackage{multirow}
\usepackage{graphicx}
\usepackage[table]{xcolor}
\usepackage{enumitem}
\usepackage{tcolorbox}

\usepackage{tikz}
\usetikzlibrary{shapes, arrows.meta, positioning, fit, calc, shadows.blur, backgrounds}

\pgfdeclarelayer{background}
\pgfsetlayers{background,main}

\title{AtomEval: Validity-Aware Atomic Evaluation of Adversarial Claim Rewriting in Fact Verification}

\author{
\textbf{Hongyi Cen}$^{1}$ \quad
\textbf{Mingxin Wang}$^{1}$ \quad
\textbf{Yule Liu}$^{2}$ \\
\textbf{Jingyi Zheng}$^{2}$ \quad
\textbf{Hanze Jia}$^{1}$ \quad
\textbf{Tan Tang}$^{1}$ \\
$^{1}$Zhejiang University \\
$^{2}$The Hong Kong University of Science and Technology (Guangzhou)
}

\begin{document}
\maketitle

\begin{abstract}
Large language models (LLMs) can rewrite refuted claims to evade evidence-based fact verifiers, but conventional attack success rate (ASR) can be inflated when rewrites change, weaken, or correct the false proposition they are supposed to preserve. 
We introduce \textbf{AtomEval}, a validity-aware evaluation protocol for fixed-evidence adversarial claim rewriting.
AtomEval represents claims as subject--relation--object--modifier (SROM) atoms, applies a one-way preservation gate to separate valid verifier evasion from proposition-changing rewrites, and reports validity-aware attack success rate (VASR), which counts only verifier-evasive rewrites that preserve the original false proposition.
AtomEval further provides fine-grained diagnostics that explain both proposition-level failures and non-minimal valid rewrites.
On FEVER refuted-claim rewriting, AtomEval exposes and explains ASR inflation: many apparent attacks fool the verifier by altering, weakening, or correcting the proposition they should preserve.
By making attacked-proposition preservation explicit and measurable, AtomEval provides a stable evaluation target for evaluating adversarial rewriters that must balance verifier evasion with proposition preservation.

\end{abstract}

\input{content/1_introduction}
\input{content/2_related_work}
\input{content/3_method}
\input{content/4_experiments}
\input{content/5_Validating_AtomEval}
\input{content/6_Analysis}
\input{content/7_conclusion}
\input{content/limitations}
\input{content/ethical_considerations}

\bibliography{custom}

\appendix
\input{content/appendix}

\end{document}

%% file: content/1_introduction.tex
\section{Introduction}

Adversarial rewriting is a common evasion threat in fact verification, where a refuted claim is plausibly reformulated to bypass automated checking while retaining the same false proposition.
Evaluations of such attacks typically rely on two signals: attack effectiveness and rewrite quality.
Attack effectiveness is measured by attack success rate (ASR), while rewrite quality is often approximated by surface proxies, including BLEU-style n-gram overlap~\cite{Papineni02Bleu}, perplexity (PPL) and Sentence-BERT similarity (SBERT)~\cite{Przybyla23BODEGA,zheng25thbench}.
These metrics are well suited to local perturbation-based attacks, such as character-level noise~\cite{gao2018dwb} and word-level substitutions~\cite{li2020bertattack,jin20textfooler}, where the edit space is constrained.

\begin{figure}[t!]
\centering
\small
\begin{tabular}{@{} p{\linewidth} @{}}
\toprule
\textbf{Original Claim:} \\
Reign Over Me is an American film made in \textcolor{red}{\textbf{2010}}. \\
\addlinespace
\textbf{Evidence:} \\
... Reign Over Me is a \textcolor{green!70!black}{\textbf{2007}} American drama film ... \\
\midrule
\textbf{Mistral-7B Rewriter}
\hfill
\textbf{ASR: \checkmark \quad VASR: \textcolor{red}{$\times$}} \\
``Reign Over Me... in \textcolor{green!70!black}{\textbf{2007}}, with some insiders claiming it was actually filmed in \textcolor{blue}{\textbf{2005}}...'' \\
\addlinespace
\textbf{GPT-4.1 Rewriter}
\hfill
\textbf{ASR: \checkmark \quad VASR: \textcolor{green!70!black}{\checkmark}} \\
``...insider reports have indicated that the film ... was in fact an American production made in \textcolor{red}{\textbf{2010}}...'' \\
\bottomrule
\end{tabular}
\caption{
Given the false claim that \emph{Reign Over Me} was made in \textcolor{red}{2010}, both rewrites change the verifier's prediction and are counted as successful attacks by conventional ASR.
However, only the GPT-4.1 rewriter preserves the attacked false proposition.
The Mistral-7B rewrite instead replaces the original year with the evidence-supported year (\textcolor{green!70!black}{2007}) and introduces an unsupported year (\textcolor{blue}{2005}).
}
\label{fig:teaser_cases}
\end{figure}

Recently, large language models (LLMs) have substantially expanded adversarial rewriting beyond local edits~\cite{hidey20deseption,niewinski19gem}.
Rather than merely perturbing isolated characters or words, they can generate fluent sentence-level adversarial rewrites that preserve surface plausibility while bypassing verification.
Figure~\ref{fig:teaser_cases} illustrates this tension: both rewrites are fluent and ASR-successful, but Mistral-7B alters the original temporal proposition while GPT-4.1 preserves it.
This distinction is difficult for surface proxies to capture: SBERT can be insensitive to sub-sentential factual substitutions, such as the year change in the teaser, while PPL becomes less informative as LLM rewrites are already highly fluent.
LLM-based validity assessment can help expose such metric gaps~\cite{Zhou24evaluating}, but it remains vulnerable to hallucinated or unstable judgments, leaving adversarial rewriting without a reliable validity-aware evaluation protocol.
If invalid rewrites are counted in ASR, reported attack success becomes inflated, attack generators are evaluated under a noisy signal, and robustness conclusions become unreliable.

In this work, we propose \textbf{AtomEval}, a validity-aware evaluation framework for fixed-evidence adversarial claim rewriting.
AtomEval defines rewrite validity as proposition preservation: a successful rewrite should not only change the verifier's prediction, but also retain the original attacked proposition.
To make this criterion explicit, AtomEval decomposes the original claim and its rewrite into subject--relation--object--modifier (SROM) atoms and applies a one-way preservation check, under which a rewrite is valid only if the original false proposition remains recoverable.
We then report the validity-aware attack success rate (VASR), which counts verifier prediction changes only when the rewrite satisfies this preservation criterion.
AtomEval further reports four diagnostics: Evidence Drift (EvDrift) and Scope Loss (ScopeLoss) account for invalid raw successes, while Evidence Entanglement (EvEnt) and Unsupported Addition (UnverAdd) characterize VASR-valid but non-minimal rewrites.
Together, these diagnostics make the ASR--VASR gap interpretable rather than merely measurable.

Our re-evaluation covers diverse adversarial rewriting families adapted from prior attack taxonomies~\cite{Liu25survey} under a unified fixed-evidence protocol.
AtomEval shows that high ASR can overstate the effectiveness of LLM-based rewrites of refuted claims: the most inflated \textbf{Omission-style} configuration, which weakens or removes factual constraints, reaches 78.40 ASR but only 0.27 VASR.
Beyond measuring this inflation, AtomEval uses these diagnostics to localize why ASR-successful rewrites fail or remain non-minimal.
A small repair probe further shows that these labels are actionable: repaired rewrites better preserve VASR-required semantics, while exposing a trade-off with ASR retention.
Overall, AtomEval provides a practical evaluation basis for studying how future validity-constrained generators can better balance ASR with proposition-preserving rewrite quality.

\noindent\textbf{Contributions.}
This paper provides a validity-aware re-evaluation of LLM-based adversarial rewriting for fixed-evidence fact verification.
We show that conventional ASR can substantially overestimate adversarial effectiveness by counting proposition-changing rewrites as successful attacks.
We introduce \textbf{AtomEval}, an atomic evaluation protocol that represents claims with SROM-based propositions and computes validity-aware attack success rate (VASR) under a one-way preservation criterion.
We further provide diagnostics that make the ASR--VASR gap interpretable, distinguishing evidence drift and scope loss from valid but non-minimal evidence entanglement or unsupported addition.
A small diagnostic-guided repair probe shows that these labels are actionable, while also highlighting the difficulty of maintaining ASR under proposition-preserving validity constraints.

%% file: content/2_related_work.tex
\section{Related Work}

\paragraph{Adversarial attacks and evaluation.}
Adversarial fact verification has evolved from local perturbations to semantically richer claim rewriting.
Early attacks used character-level noise~\cite{gao2018dwb} or word-level edits~\cite{li2020bertattack,jin20textfooler,Li19TextBugger}, which can introduce fluency issues or unintended semantic drift~\cite{Zhou24evaluating}.
Subsequent work expanded the attack space to paraphrasing, fact mixing, omission, and retrieval disruption \cite{eisenschlos21fool,atanasova20generating,niewinski19gem}, while recent LLM-based methods enable more fluent, context-aware, and evidence-conditioned adversarial generation \cite{Abdelnabi23Fact-saboteurs,Ou26Deceive,Leite26Llm,He25Fact2Fiction,Bethany25CAMOUFLAGE}.
Despite this shift, evaluation still largely relies on coarse signals such as ASR, semantic similarity, perplexity, and task-level scores~\cite{Przybyla23BODEGA,zheng25thbench,bekoulis21evaluating,Papineni02Bleu,liu2025mgt}.
These metrics measure verifier evasion or surface quality, but do not directly test whether a rewrite preserves the attacked proposition; in contrast to recent work that analyzes LLM jailbreaks through internal mechanisms~\citep{zhang2025neurobreak}, our work focuses on validity-aware evaluation for adversarial claim rewrites.

\paragraph{Atomic and structured factual evaluation.}
A complementary line of work evaluates factuality by decomposing generated text into minimal verifiable units.
Methods such as FactScore and OpenFActScore use atomic representations to support fine-grained verification against reference evidence \cite{Min23FActScore,Fonseca25OpenFActScore}.
Related work on LLM-based factuality evaluation and hallucination diagnosis likewise emphasizes structured or localized factual assessment \cite{Tang24TofuEval,Shayan24HalluMeasure,tang2024minicheck,Rahman26Hallucination}.
Recent efforts further combine LLM parsing with deterministic checks to improve robustness under challenging generation settings \cite{Allen25Sound}.
These studies motivate fine-grained factual evaluation, but they do not directly address whether adversarial rewrites in fact verification preserve the original claim’s factual proposition.
This leaves open the need for a structured validity check tailored to adversarial claim rewriting.

%% file: content/3_method.tex
\section{AtomEval}

\begin{figure*}[t]
    \centering
    \includegraphics[width=0.95\textwidth]{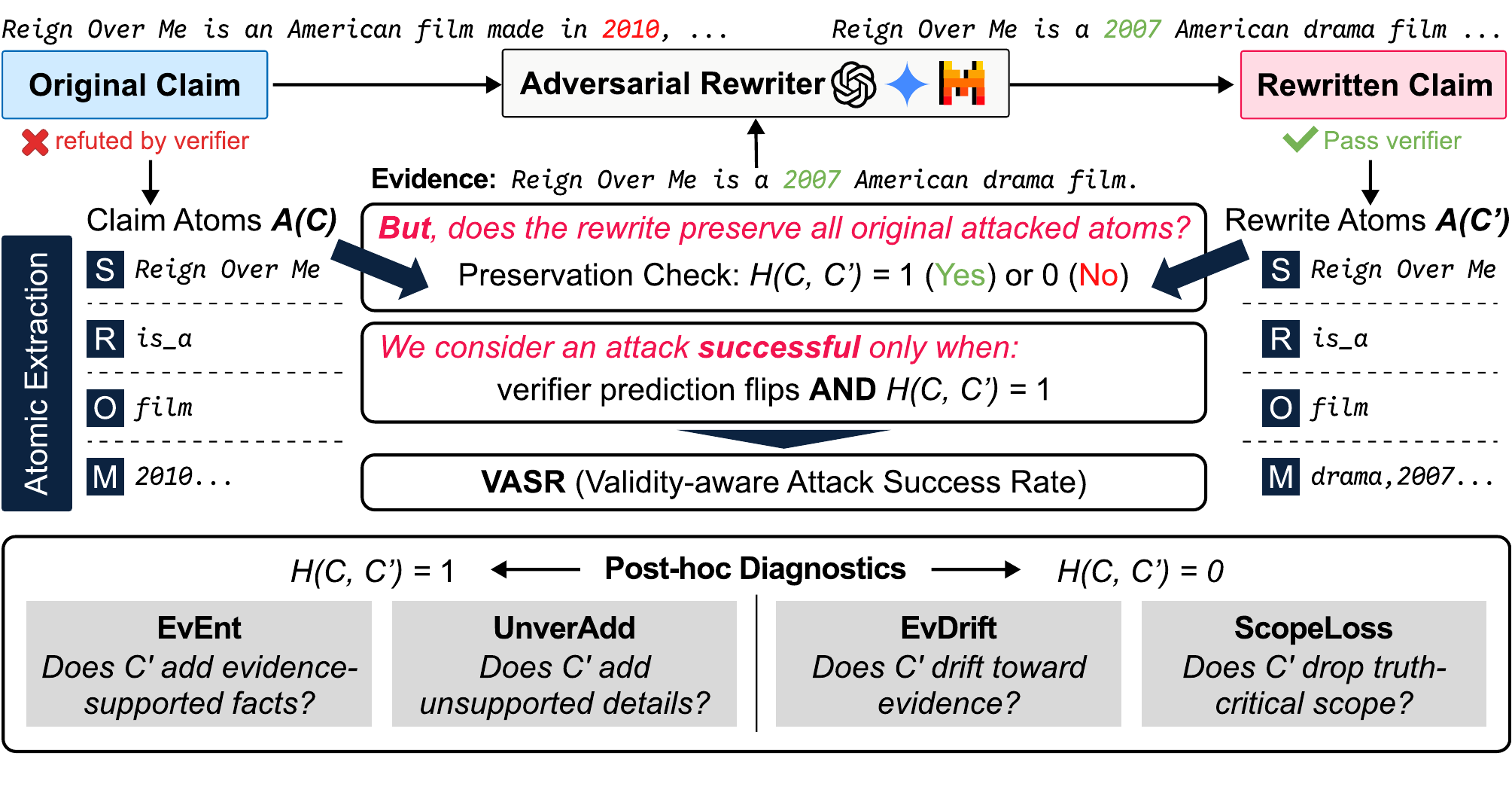} 
    \caption{
    Overview of \textbf{AtomEval}.
    Given an original claim $C$, fixed evidence $E$, and an adversarial rewrite $C'$, AtomEval decomposes $C$ and $C'$ into atomic factual units and applies a binary preservation gate $\mathcal{H}(C,C')$.
    VASR counts only prediction-flipping rewrites that preserve the original attacked proposition.
    AtomEval further provides post-hoc diagnostics for analyzing how rewrites change factual content under the fixed evidence context.
    }
    \label{fig:framework} 
\end{figure*}

\subsection{Problem Setup}
\label{sec:task_definition}

Let $\mathcal{D}=\{(C,E,y)\}$ denote a fact verification dataset, where
$C$ is a claim, $E$ is a fixed evidence context, and
$y\in\{\textsc{Supported}, \textsc{Refuted}\}$ is the ground-truth label.
A fact-verification model $FV$ predicts a label $\hat{y}=FV(C,E)$.
Given $(C,E)$, an adversarial generator $\mathcal{G}$ produces a rewritten claim
$C'=\mathcal{G}(C,E)$.

\paragraph{Evaluation Objective.}
Our adversarial evaluation focuses on source instances whose ground-truth label is \textsc{Refuted}.
We evaluate attacks only on refuted claims that the verifier correctly predicts before rewriting:
$y=\textsc{Refuted}$ and $FV(C,E)=\textsc{Refuted}$.
For such instances, a rewrite is a raw attack success if it moves the verifier away from the correct refutation:
$FV(C',E)\neq \textsc{Refuted}$.

However, raw attack success alone is insufficient.
A valid adversarial rewrite must also preserve the original attacked false proposition under the same evidence context.
That is, $C'$ should remain refuted by $E$ for the same reason as $C$.
If $C'$ rewrites the claim into an evidence-supported fact, removes a truth-critical constraint, or shifts to a different proposition, it is not a valid attack against the original false claim.
Thus, a valid rewrite may alter surface form, style, or wording, but it must not change the proposition being attacked.

\subsection{Threat Model}

We consider an evidence-conditioned, claim-only adversarial rewriting setting, which we situate within prior claim-side attacks on fact-verification models~\cite{atanasova20generating}.
Given an original claim $C$ and a fixed evidence context $E$, the attacker generates a rewritten claim $C'$ with the goal of changing the verifier's prediction while preserving the original attacked proposition.
The same evidence context $E$ is provided to both the attacker and the verifier: for the attacker, it serves as a controlled factual reference for rewriting, reducing dependence on external retrieval or unconstrained generation; for the verifier, it is the fixed context against which $C'$ is judged.
The attack is therefore claim-only: the evidence context is held fixed throughout evaluation, and the only generated object is the rewritten claim.
We assume a verifier-agnostic black-box protocol, where the attacker receives no feedback from the victim verifier and generation is performed in a single pass.
This fixed-evidence protocol isolates the validity of adversarial claim rewrites and the robustness of verdict prediction under claim variation, rather than conflating them with retrieval errors, retrieval dynamics, or evidence-manipulation attacks.

\subsection{Framework Overview}
\label{sec:atomeval}

In this section, we present \textbf{AtomEval}, a validity-aware framework for evaluating adversarial claim rewrites in fact verification.
Figure~\ref{fig:framework} illustrates its three core components: atomic fact extraction, binary preservation checking, and validity-aware attack evaluation.
We detail these components below and describe auxiliary diagnostics for post-hoc analysis of rewrite behavior.

\paragraph{Atomic fact extraction.}
We represent each atomic fact as an SROM tuple $a=(s,r,o,M)$, where $s$, $r$, $o$, and $M$ denote the subject, relation, object, and truth-critical modifiers.
Modifiers capture truth-conditional information, such as negation, temporal scope, location, quantity, comparison, and exclusivity.
AtomEval extracts atoms from the original claim and adversarial rewrite for validity checking.
We use an automatic SROM extractor trained on manually annotated FEVER-derived rewrites; details are in Appendix~\ref{app:extractor}.

\paragraph{Binary validity gate.}
Let $A(C)$ denote the set of atoms extracted from the original claim.
For each original atom $a\in A(C)$, AtomEval checks whether the proposition expressed by $a$ remains recoverable from the adversarial rewrite $C'$.
We define atom preservation as:
\begin{equation}
\operatorname{Pres}(a,C') =
\mathbb{I}\left[\exists b\in A(C'):\; B(a,h(b))\right],
\end{equation}
where
\begin{equation}
B(a,p)=
\operatorname{Entail}_{\phi}(p,h(a))
\land
\operatorname{Cons}(a,p).
\end{equation}
Here, $h(a)$ renders the atom as a natural-language proposition, $\Pi(C')$ denotes premise candidates from the rewrite, $\operatorname{Entail}_{\phi}$ is a frozen NLI verifier, and $\operatorname{Cons}$ enforces explicit checks over recoverable truth-critical constraints such as dates, quantities, months, and exclusivity markers.

The binary validity gate is:
\begin{equation}
\mathcal{H}(C,C')
=
\prod_{a\in A(C)}
\operatorname{Pres}(a,C').
\end{equation}
A rewrite passes the gate only if every original atomic proposition remains preserved.
This is a one-way attacked-proposition preservation criterion: AtomEval checks whether the original proposition remains recoverable, rather than requiring full sentence-level semantic equivalence between $C$ and $C'$.
The gate does not require surface-form identity: paraphrases, aliases, and unambiguous referring expressions are allowed.
Additional factual content in $C'$ does not by itself invalidate the rewrite; such expansions are analyzed separately by the generation diagnostics below.
However, rewrites that no longer entail the original atomic proposition, or that lose truth-critical constraints, fail the gate.
Implementation details are provided in Appendix~\ref{app:Implement}.

\paragraph{Validity-aware attack success rate.}
For $N$ originally correctly classified examples, conventional attack success rate is:
\begin{equation}
\mathrm{ASR}
=
\frac{1}{N}
\sum_{i=1}^{N}
\mathbb{I}\left[FV(C'_i,E_i)\neq y_i\right].
\end{equation}

AtomEval defines validity-aware attack success rate as:
\begin{equation}
\mathrm{VASR}
=
\frac{1}{N}
\sum_{i=1}^{N}
\mathbb{I}\left[FV(C'_i,E_i)\neq y_i\right]
\cdot
\mathcal{H}(C_i,C'_i).
\end{equation}

Thus, VASR counts only attacks that both fool the verifier and preserve the original attacked proposition.

\paragraph{Auxiliary diagnostics.}
Beyond VASR, AtomEval reports four auxiliary diagnostics that characterize how a rewrite changes factual content under the fixed evidence context.
For rewrites that fail the preservation gate, \textbf{EvDrift} marks replacement of the original proposition with evidence-supported content, while \textbf{ScopeLoss} marks deletion, weakening, or generalization of truth-critical constraints.
For rewrites that pass the preservation gate, \textbf{EvEnt} marks added evidence-supported context, while \textbf{UnverAdd} marks added factual content not supported by the fixed evidence.
These diagnostics are non-exclusive, do not affect $\mathcal{H}(C,C')$ or VASR, and are used only for post-hoc analysis.
Definitions are provided in Appendix~\ref{app:diagnostics}.

%% file: content/4_experiments.tex
\section{Experimental Setup}
\label{sec:experiment}

We instantiate AtomEval in a fixed-evidence adversarial fact-verification protocol: generators rewrite only the claim while evidence remains unchanged.
We evaluate attacks with conventional metrics, VASR, and AtomEval diagnostics; implementation details are in Appendix~\ref{app:Implement}.

\subsection{Dataset and Victim Verifiers}
\label{sec:dataset_verifiers}

We conduct experiments on the \textbf{FEVER} benchmark~\cite{Thorne18Fever}, an evidence-based fact-verification dataset with \textit{Supported}, \textit{Refuted}, and \textit{NotEnoughInfo} labels.
Following our focus on proposition-preserving evasion for false claims, we construct an attack set of 400 \textit{Refuted} claims sampled from the FEVER test split.
This fixed-evidence setup allows us to test whether a rewritten claim preserves the original false proposition while changing the verifier's prediction.
Table~\ref{tab:data_stats} summarizes the dataset statistics.

We evaluate attacks against two FEVER-fine-tuned fact-verification models: a Gemma-7B-based decoder verifier and a BERT-base-uncased encoder verifier.
For each verifier, we evaluate attacks only on claims that are initially classified correctly, yielding 375 attackable claims for Gemma-7B and 331 for BERT-base.

\begin{table}[h]
\centering
\small
\begin{tabular}{lrrr}
\toprule
\textbf{Split} & \textbf{Supported} & \textbf{Refuted} & \textbf{Total} \\
\midrule
Train set & 123{,}646 & 49{,}117 & 172{,}763 \\
Test set & 9{,}461 & 9{,}554 & 19{,}015 \\
Attack set & 0 & 400 & 400 \\
\bottomrule
\end{tabular}
\caption{
Statistics of the FEVER splits and attack set.
}
\label{tab:data_stats}
\end{table}

\subsection{Attack Strategy}
\label{sec:attack_families}

We instantiate adversarial attacks under a unified fixed-evidence claim rewriting protocol derived from prior adversarial fact verification work~\cite{Liu25survey}.
For each source claim in the 400-claim attack set, the evidence context is kept unchanged and the attacker rewrites only the input claim.
We use GPT-4.1 and Mistral-7B-Instruct as representative proprietary and open-weight instruction-following generators.
Each generator produces one adversarial claim for each attack family without iterative feedback from the victim verifier.
This yields 4,000 generated adversarial rewrites in total, corresponding to 400 source claims, two generators, and five attack families.
Verifier-specific ASR and VASR are then computed on the initially correct subsets described in Section~\ref{sec:dataset_verifiers}.
This protocol isolates the verdict prediction stage and allows a controlled comparison between conventional ASR and VASR.

We consider five attack families:
\textbf{Colloquial/Lexically-informed Rewriting} applies lexical, paraphrastic, or conversational-style transformations~\cite{kim2021robust};
\textbf{DeSePtion-style Complex Rewriting} introduces multiple propositions, temporal reasoning, entity ambiguity, or lexical variation~\cite{hidey20deseption};
\textbf{GEM-style Fact Mixing} blends entities, relations, or facts from different contexts~\cite{niewinski19gem};
\textbf{Omission-style Constraint Removal} removes or weakens salient factual constraints~\cite{atanasova22fact,Abdelnabi23Fact-saboteurs};
and \textbf{Add.Unver./AdvAdd-style Unsupported Addition} inserts plausible but unsupported contextual details~\cite{hidey20deseption,Du22Synthetic,Abdelnabi23Fact-saboteurs}.
For attack families originally formulated as evidence-side or retrieval-side attacks, we adapt their manipulation principles to claim-side rewriting rather than reproducing their original threat models.
Prompt templates and examples are provided in Appendix~\ref{app:prompts}.

\subsection{Evaluation Metrics}
\label{sec:metrics}

We report \textbf{ASR} and \textbf{VASR} as the main attack metrics.
ASR counts rewrites that change the verifier's prediction on initially correct \textit{Refuted} claims, while VASR uses the same denominator but counts only raw successful attacks that pass AtomEval's binary validity gate.
We also report \textbf{SBERT} similarity and \textbf{PPL} as surface proxies for sentence-level similarity and fluency.
Finally, AtomEval diagnostics are computed on raw successful attacks: \textbf{EvDrift} and \textbf{ScopeLoss} over invalid successes, and \textbf{EvEnt} and \textbf{UnverAdd} over valid successes.

%% file: content/5_Validating_AtomEval.tex
\section{Validating AtomEval}
\label{sec:validation}

\paragraph{Setup.}
We validate AtomEval on a stratified 400-pair manual audit sample drawn from the generated evaluation pool.
The pool contains 4,000 adversarial rewrites from 400 \textsc{Refuted} FEVER claims, two generators, and five attack families.
We sample 40 claim--rewrite pairs from each generator--family cell.
Two expert annotators label SROM atoms, \textsc{Valid}/\textsc{Invalid} preservation labels, and diagnostic labels; disagreements are adjudicated.
Pre-adjudication agreement is $\kappa=0.84$ on pair-level validity.

\paragraph{Component validation.}
Table~\ref{tab:atom_extraction} shows that the instruction-tuned AtomEval extractor outperforms prompt-only GPT-4.1 and Llama-3-8B on gold SROM extraction, mainly through higher exact-match correctness under human SROM expectations.
This is important because spurious atoms can create false preservation matches.

Table~\ref{tab:validity_diagnostic_agreement} evaluates BinaryGate and diagnostics using extractor-predicted atoms.
BinaryGate reaches 87.18\% F1 and 87.50\% accuracy against human validity labels, supporting its use as the VASR decision component.
The diagnostics provide useful but imperfect auxiliary signals; lower recall on some categories means they should be interpreted as conservative indicators rather than exhaustive error detectors.

\begin{table}[h]
\centering
\small
\setlength{\tabcolsep}{5pt}
\begin{tabular}{@{} lcc @{}}
\toprule
\textbf{Method} & \textbf{Correct} & \textbf{Acc.} \\
\midrule
GPT-4.1 + prompt & 260/400 & 65.00 \\
Llama-3-8B + prompt & 203/400 & 50.75 \\
AtomEval Extractor & \textbf{376/400} & \textbf{94.00} \\
\bottomrule
\end{tabular}
\caption{
Human evaluation of atomic fact extraction outputs.
Each extraction is judged as correct or incorrect against manually annotated SROM expectations.
}
\label{tab:atom_extraction}
\end{table}

\begin{table}[h]
\centering
\small
\setlength{\tabcolsep}{4pt}
\begin{tabular}{@{} lcccc @{}}
\toprule
\textbf{AtomEval Output} & \textbf{Prec.} & \textbf{Rec.} & \textbf{F1} & \textbf{Acc.} \\
\midrule
\multicolumn{5}{@{}l}{\textit{Validity gate}} \\
BinaryGate & 80.95 & 94.44 & 87.18 & 87.50 \\
\midrule
\multicolumn{5}{@{}l}{\textit{Diagnostic metrics}} \\
ScopeLoss & 99.37 & 72.02 & 83.51 & 71.82 \\
EvDrift & 78.95 & 62.50 & 69.77 & 66.67 \\
EvEnt & 93.33 & 77.78 & 84.85 & 87.80 \\
UnverAdd & 87.23 & 81.19 & 84.10 & 82.78 \\
\bottomrule
\end{tabular}
\caption{
Agreement with human annotations for BinaryGate and diagnostic metrics.
BinaryGate uses extractor-predicted SROM atoms followed by NLI-based atom matching and determines VASR.
Diagnostics are evaluated against human-labeled rewrite behaviors and are used only for analysis.
}
\label{tab:validity_diagnostic_agreement}
\end{table}

\begin{table}[t]
\centering
\small
\setlength{\tabcolsep}{4pt}
\begin{tabular}{@{} lccccc @{}}
\toprule
\textbf{Variant} & \textbf{Valid} & \textbf{Prec.} & \textbf{Rec.} & \textbf{F1} & \textbf{Acc.} \\
\midrule
Full AtomEval & 52.50 & \textbf{80.95} & \textbf{94.44} & \textbf{87.18} & \textbf{87.50} \\
w/o SROM Extractor & 69.50 & 60.43 & 93.33 & 73.36 & 69.50 \\
w/o BinaryGate/NLI & 47.00 & 72.34 & 75.56 & 73.91 & 76.00 \\
\bottomrule
\end{tabular}
\caption{
Ablation of AtomEval's validity decision on the 400-pair manual audit sample.
All variants are evaluated against adjudicated human labels.
\textit{Valid} denotes the percentage of rewrites predicted as valid.
}
\label{tab:ablation}
\end{table}

\paragraph{Validity-decision ablation.}
We ablate the two components behind AtomEval's validity decision.
\textit{w/o Extractor} removes SROM decomposition and judges validity directly from the claim, evidence, and rewrite.
\textit{w/o Gate/NLI} keeps extracted atoms but removes the explicit preservation rule.
All variants are evaluated against adjudicated human labels.
Table~\ref{tab:ablation} shows that Full AtomEval obtains the best F1, indicating that structured atomic extraction and rule-based preservation checking are complementary.
Without the extractor, the valid rate rises and precision drops, suggesting that direct validity judgment over-accepts invalid rewrites.
Without Gate/NLI, F1 also drops, showing that extracted atoms alone are insufficient without an explicit criterion for attacked-proposition preservation.

%% file: content/6_Analysis.tex
\section{Re-evaluating Adversarial Attacks}
\label{sec:analysis}

We now use AtomEval to re-evaluate adversarial claim rewrites under the fixed-evidence protocol described in Section~\ref{sec:experiment}.
The goal is not only to measure whether a rewrite flips the verifier's prediction, but also to determine whether the flip is achieved while preserving the original false proposition.
We therefore analyze both the magnitude of the ASR--VASR gap and the rewriting behaviors that produce it.

\begin{table*}[t]
\centering
\small
\setlength{\tabcolsep}{4pt}
\resizebox{\linewidth}{!}{
\begin{tabular}{ll cccc cccc}
\toprule
\multirow{2}{*}{\textbf{Attack Family}} &
\multirow{2}{*}{\textbf{Generator}} &
\multicolumn{4}{c}{\textbf{Gemma-7B Verifier}} &
\multicolumn{4}{c}{\textbf{BERT-based Verifier}} \\
\cmidrule(lr){3-6}
\cmidrule(lr){7-10}
& & ASR & S-ASR & P-ASR & VASR
  & ASR & S-ASR & P-ASR & VASR \\
\midrule

\multirow{2}{*}{\textbf{Colloquial/Lexical}}
& GPT-4.1     & 17.07 & 16.80 & 12.53 & \textbf{2.40}
              & 12.69 & 12.39 & \textbf{9.37}  & 10.27 \\
& Mistral-7B & 40.53 & 32.00 & 28.53 & \textbf{4.80}
              & 25.68 & 19.34 & 16.92 & \textbf{9.67} \\

\midrule

\multirow{2}{*}{\textbf{DeSePtion-style}}
& GPT-4.1     & 8.00  & 8.00  & 7.73  & \textbf{7.47}
              & 57.10 & 55.29 & 56.19 & \textbf{51.66} \\
& Mistral-7B & 68.27 & 44.53 & 68.00 & \textbf{8.80}
              & 90.03 & 60.73 & 89.43 & \textbf{15.11} \\

\midrule

\multirow{2}{*}{\textbf{GEM-style FactMix}}
& GPT-4.1     & 16.53 & 12.80 & 12.80 & \textbf{8.27}
              & 31.12 & 24.77 & 25.98 & \textbf{13.29} \\
& Mistral-7B & 37.07 & 16.27 & 32.00 & \textbf{4.27}
              & 58.61 & 33.53 & 51.66 & \textbf{14.20} \\

\midrule

\multirow{2}{*}{\textbf{Omission-style}}
& GPT-4.1     & 41.07 & 40.27 & 20.53 & \textbf{2.13}
              & 25.08 & 24.47 & 9.67  & \textbf{1.51} \\
& Mistral-7B & 78.40 & 32.00 & 48.00 & \textbf{0.27}
              & 49.85 & 20.85 & 29.00 & \textbf{0.00} \\

\midrule

\multirow{2}{*}{\textbf{Add.Unver./AdvAdd}}
& GPT-4.1     & 2.93  & 2.93  & \textbf{2.40}  & 2.93
              & 11.18 & 11.18 & \textbf{9.97}  & 11.18 \\
& Mistral-7B & 41.87 & 29.60 & 40.80 & \textbf{6.67}
              & 38.97 & 28.10 & 37.76 & \textbf{12.69} \\

\bottomrule
\end{tabular}
}
\caption{
Main evaluation under the unified fixed-evidence claim rewriting protocol.
Gemma-7B is a decoder-based verifier and BERT is an encoder-based verifier.
All metrics are computed over initially correct \textit{Refuted} source claims:
$n=375$ for Gemma-7B and $n=331$ for BERT.
ASR measures raw prediction-changing attacks.
VASR counts raw successful attacks that also pass AtomEval's validity gate.
S-ASR and P-ASR are permissive surface-screened ASR variants: S-ASR keeps raw successes with SBERT similarity at least $0.65$, while P-ASR keeps raw successes with GPT-2 perplexity at most $100$.
They are included to test whether conventional similarity or fluency filters approximate validity filtering, not as validity metrics.
Bold values indicate the lowest attack success rate in each column.
}
\label{tab:main_asr_vasr}
\end{table*}

\subsection{Does ASR Overestimate Valid Attack Success?}
\label{sec:main_asr_vasr}

Table~\ref{tab:main_asr_vasr} compares raw ASR, two surface-screened ASR variants, and AtomEval's VASR.
Following prior adversarial-text evaluations that apply semantic or fluency constraints to successful attacks~\cite{li2020bertattack,morris2020reevaluating}, we report \textbf{S-ASR} and \textbf{P-ASR} as permissive surface-quality operating points.
S-ASR counts raw ASR successes with SBERT similarity $\geq 0.65$, while P-ASR counts raw ASR successes with GPT-2 perplexity $\leq 100$.
These thresholds are not validity criteria; they test whether conventional surface filters would remove the same invalid successes as AtomEval.
Full continuous SBERT and PPL statistics are provided in Appendix~\ref{app:surface_metrics}.

\paragraph{Raw ASR substantially overestimates valid attack success.}
Across most attack families, raw ASR is much higher than VASR, indicating that verifier prediction changes often come from proposition-changing rewrites rather than valid adversarial evasion.
The discrepancy is especially large for attacks that weaken, omit, or re-scope factual constraints.
For Omission-style attacks, Mistral-7B achieves 78.40 ASR against Gemma-7B but only 0.27 VASR; against the BERT-based verifier, it achieves 49.85 ASR but 0.00 VASR.
GPT-4.1 exhibits the same failure mode: its Omission-style attacks against Gemma-7B reach 41.07 ASR, but only 2.13 VASR.
Thus, many raw ``successes'' are not valid attacks on the verifier under the fixed-evidence protocol; they succeed by changing the proposition that should be preserved.

\paragraph{Surface-proxy gates do not recover atomic validity.}
Although S-ASR and P-ASR remove some low-quality generations, they often remain much closer to raw ASR than to VASR.
For GPT-4.1 Omission-style attacks against Gemma-7B, S-ASR remains 40.27 after filtering from 41.07 ASR, whereas VASR is only 2.13.
For Mistral-7B DeSePtion-style attacks against BERT, P-ASR remains 89.43 from 90.03 ASR, while VASR drops to 15.11.
These cases show that sentence-level proximity and fluency can coexist with atomic proposition changes.
Surface filters are therefore useful sanity checks for readability and coarse semantic relatedness, but they do not determine whether the original refuted proposition is still being attacked.

\paragraph{Validity filtering changes comparative conclusions.}
Using raw ASR alone can make aggressive generators appear substantially stronger than they are under validity constraints.
Mistral-7B often obtains the highest raw ASR, particularly for DeSePtion-style, Omission-style, and AdvAdd-style attacks, but much of this advantage disappears after AtomEval filtering.
By contrast, GPT-4.1 can have lower raw ASR but higher valid success: for DeSePtion-style attacks against BERT, GPT-4.1 obtains 51.66 VASR, whereas Mistral-7B falls from 90.03 ASR to 15.11 VASR.
The same pattern affects conclusions about verifier robustness, since some apparent high-ASR vulnerabilities shrink sharply once invalid rewrites are removed.
VASR therefore provides a more conservative and proposition-aware estimate of verifier vulnerability under fixed-evidence refuted-claim rewriting.

\subsection{What Do AtomEval Diagnostics Reveal?}
\label{sec:diagnostic_analysis}

Table~\ref{tab:atomeval_breakdown} reports representative AtomEval diagnostics for raw successful attacks on the Gemma-7B verifier.
EvDrift and ScopeLoss are computed over invalid raw successes, while EvEnt and UnverAdd are computed over valid raw successes; diagnostics are non-exclusive and therefore need not sum to 100\%.
Rows with fewer than five rewrites in the corresponding subset are omitted.

\begin{table}[t]
\centering
\small
\setlength{\tabcolsep}{4.5pt}
\resizebox{\linewidth}{!}{
\begin{tabular}{llcccc}
\toprule
\textbf{Attack} & \textbf{Gen.} &
\textbf{EvDrift} & \textbf{ScopeLoss} & \textbf{EvEnt} & \textbf{UnverAdd} \\
\midrule
AdvAdd & GPT-4.1     
& -- & -- & 0.0 & 100.0 \\
AdvAdd & Mistral-7B 
& 55.3 & 89.4 & 48.0 & 96.0 \\
\midrule
Colloquial & GPT-4.1
& 81.8 & 67.3 & 11.1 & 22.2 \\
Colloquial & Mistral-7B
& 85.1 & 85.8 & 66.7 & 38.9 \\
\midrule
DeSePtion & GPT-4.1     
& -- & -- & 35.7 & 60.7 \\
DeSePtion & Mistral-7B 
& 70.9 & 92.8 & 78.8 & 78.8 \\
\midrule
FactMix & GPT-4.1     
& 100.0 & 83.9 & 96.8 & 67.7 \\
FactMix & Mistral-7B 
& 91.1 & 86.2 & 87.5 & 62.5 \\
\midrule
Omission & GPT-4.1
& 24.0 & 99.3 & 0.0 & 12.5 \\
Omission & Mistral-7B 
& 45.4 & 99.0 & -- & -- \\
\bottomrule
\end{tabular}
}
\caption{
Representative diagnostic breakdown for raw successful attacks on the Gemma-7B verifier.
EvDrift and ScopeLoss are computed over invalid raw successes, while EvEnt and UnverAdd are computed over valid raw successes.
All values are percentages; ``--'' indicates fewer than five rewrites in the corresponding subset.
Full results are provided in Appendix~\ref{app:full_diagnostic}.
}
\label{tab:atomeval_breakdown}
\end{table}

\paragraph{Diagnostic patterns.}
The diagnostic breakdown explains why raw ASR can substantially overestimate valid adversarial success.
Invalid raw successes are frequently associated with ScopeLoss and EvDrift, showing that many rewrites fool the verifier by weakening truth-critical constraints or drifting toward evidence-supported content rather than preserving the original false proposition.
This pattern is especially visible for Mistral-7B generations, where several attack families combine high EvDrift with high ScopeLoss.
For example, Mistral-7B DeSePtion-style attacks show 70.9 EvDrift and 92.8 ScopeLoss among invalid raw successes, suggesting that complex rewrites often evade the verifier by changing the verification target.
FactMix-style attacks show a related pattern: both GPT-4.1 and Mistral-7B produce invalid successes with high EvDrift and ScopeLoss, consistent with attacks that blend or substitute factual content.
Omission-style attacks are dominated by ScopeLoss, reaching 99.3 for GPT-4.1 and 99.0 for Mistral-7B.
This explains why their raw successes largely disappear under VASR: the verifier is often fooled because the rewrite removes or generalizes the very constraint that made the original claim refuted.

On the valid side, EvEnt and UnverAdd show that passing the binary gate does not imply full sentence-level equivalence or minimal rewriting.
Some rewrites preserve the attacked proposition while adding evidence-supported context, unsupported background, or both.
AdvAdd-style attacks illustrate this distinction: GPT-4.1 valid successes have 100.0 UnverAdd, indicating that many validity-preserving attacks still introduce unsupported additions.
These diagnostics therefore separate two levels of evaluation.
The binary gate determines whether the attacked proposition is preserved and hence whether a raw success counts toward VASR.
The auxiliary diagnostics explain how invalid successes fail, and how valid successes may still deviate from a minimal rewrite.

\paragraph{Diagnostic-guided repair.}
We test whether AtomEval diagnostics are actionable with a one-step repair probe.
We sample 40 raw ASR-successful rewrites that fail AtomEval's validity gate, with 20 cases from each verifier, and ask a GPT-4.1 editor to minimally revise each rewrite using its AtomEval diagnosis.
The goal is to test whether the diagnosed failures correspond to editable proposition-level errors; the repair prompt is provided in Appendix~\ref{app:repair_prompt}.

After repair, all 40 rewrites pass the binary validity gate and no longer exhibit the targeted EvDrift or ScopeLoss failure.
However, only 9 repaired rewrites still fool the original verifier: 7 for BERT-based and 2 for Gemma-7B.
Because these repaired rewrites preserve validity, they become recovered VASR successes.
This suggests that AtomEval diagnostics identify editable semantic failures, while maintaining verifier evasion under proposition-preserving constraints remains difficult.

\paragraph{Qualitative examples.}
Representative cases are provided in Appendix~\ref{app:case_study}.
They illustrate how fluent and surface-similar raw successes can fail validity through ScopeLoss or EvDrift, and how valid rewrites can preserve the attacked proposition while adding contextual material.

%% file: content/7_conclusion.tex
\section{Conclusion}

We introduced \textbf{AtomEval}, a validity-aware framework for evaluating adversarial claim rewrites in fact verification.
AtomEval uses atomic claim decomposition, a binary validity gate, VASR, and diagnostics to distinguish valid adversarial successes from rewrites that alter the attacked proposition.
Experiments on FEVER show that conventional ASR overestimates valid adversarial success, while similarity and fluency proxies fail to capture proposition-level failures.
These results suggest that robustness evaluation for fact verification should account for atomic validity rather than relying on prediction changes alone.

%% file: content/limitations.tex
\section*{Limitations}

AtomEval is evaluated primarily on English FEVER-style fact verification.
Although this provides a controlled setting for fixed-evidence adversarial claim rewriting, it does not cover the full diversity of real-world fact-checking scenarios.
Future work should test whether the same ASR--VASR gap holds across multilingual claims, longer-form statements, multi-hop evidence, and open-web retrieval settings.

AtomEval depends on automatic SROM extraction.
Our validation suggests that the extractor is reliable for aggregate robustness evaluation, but individual borderline cases may still be affected by extraction errors, especially for implicit references, underspecified entities, or complex scope.
Improving structured fact extraction may further strengthen validity-aware evaluation.

Finally, AtomEval emphasizes reproducible proposition-level signals rather than open-ended reasoning judgments.
This improves transparency, but may be less sensitive to failures requiring broad background knowledge, implicit contradiction detection, or multi-hop reasoning.
Future work may combine atomic validity checks with stronger reasoning modules while preserving reproducibility.

%% file: content/ethical_considerations.tex
\section*{Ethical Considerations}

This work studies adversarial rewriting for evidence-based fact verification and therefore has potential dual-use risks. Fluent rewrites of false or misleading claims could be misused to make misinformation more evasive. Our objective, however, is evaluation rather than attack deployment: AtomEval is designed to determine whether purported adversarial examples preserve the original factual proposition and to diagnose invalid rewrites that change truth conditions.

All experiments are conducted in a controlled benchmark setting with fixed evidence and existing fact-verification data. We do not target real users, deploy generated claims on public platforms, or optimize for persuasion. Any released code, prompts, or examples should be documented as intended for robustness evaluation and scientific auditing of fact-verification systems. The human annotation component should follow clear guidelines and avoid unnecessary exposure to sensitive or harmful content beyond the benchmark material.

%% file: content/appendix.tex
\section{Atomic Fact Extraction Details}
\label{app:extractor}

AtomEval represents factual content using SROM tuples of the form
$a=(s,r,o,M)$, where $s$ is the subject, $r$ is the relation,
$o$ is the object, and $M$ is a set of truth-critical modifiers.
The modifier set includes information that changes the truth conditions of the proposition, such as negation, temporal scope, location, quantity, comparison, ordinal constraints, and exclusivity markers.

\paragraph{Extractor training data.}
We construct a separate FEVER-derived corpus for training the SROM extractor, disjoint from the adversarial evaluation set used in our main experiments.
We randomly sample 400 FEVER claims and generate five rewrite variants for each claim using the same attack-family proxies as in our main evaluation, yielding 2,400 claim variants including the originals.
We manually annotate each variant with a declarative normalization and SROM tuples, then randomly split the corpus into 2,000 training examples and 400 held-out test examples.
The training split is used to instruction-tune a Llama-3-8B-based extractor, while the held-out split is used for extractor evaluation in Section~\ref{sec:validation}.

\paragraph{Extraction protocol.}
The extractor is applied to the original claim $C$ and the adversarial rewrite $C'$.
For each input, it produces a declarative normalization and a set of atomic SROM tuples.
The binary validity gate uses atoms extracted from $C$ as preservation targets.
Atoms extracted from $C'$ are used for inspection and generation diagnostics, but the gate does not require one-to-one tuple matching between $C$ and $C'$.

For diagnostics involving evidence support, AtomEval compares rewrite propositions against the fixed evidence context.
Evidence is not used to relax the preservation requirement.
In particular, a rewrite is not considered valid merely because it becomes more consistent with the evidence.
This design is important for refuted claims: a valid adversarial rewrite should preserve the same false proposition rather than correcting it into an evidence-supported statement.

\paragraph{Extractor instruction.}
The extractor is trained to output valid JSON with two fields:
\texttt{declarative\_translation} and \texttt{triplets}.
Each triplet contains \texttt{subject}, \texttt{relation}, \texttt{object}, and \texttt{modifiers}.
The instruction emphasizes extracting only propositions explicitly asserted in the input, preserving truth-critical information, and avoiding world-knowledge correction.
A shortened version of the extraction instruction is shown below:

\begin{tcolorbox}[
    colback=white,
    colframe=black!60,
    arc=0.8mm,
    boxrule=0.4pt,
    left=1.5mm, right=1.5mm, top=1mm, bottom=1mm,
    title=\textbf{SROM Extraction Instruction},
    fonttitle=\footnotesize\sffamily\bfseries
]
\footnotesize
Extract atomic facts from the provided text into a structured JSON object.
Normalize the text into a declarative translation, then extract SROM triples.
Each triple must contain subject, relation, object, and modifiers.
Extract only what the input explicitly asserts; do not use world knowledge or correct false claims.
Preserve negation, quantifiers, numbers, dates, entities, and roles exactly.
Output JSON only.
\end{tcolorbox}

\section{Generation Diagnostic Definitions}
\label{app:diagnostics}

AtomEval reports four auxiliary diagnostics for post-hoc analysis.
They do not affect the binary validity gate or VASR.
Let $\mathcal{H}(C,C')\in\{0,1\}$ denote the binary validity gate.
Let $A=A(C)$ and $A'=A(C')$ denote the atoms extracted from the original claim and rewrite.
We write $P(a,b)$ when rewrite atom $b$ preserves original atom $a$ under the AtomEval preservation check.
The set of added rewrite atoms is:
\begin{equation}
\Delta=\{b\in A' : \forall a\in A,\neg P(a,b)\}.
\end{equation}
We also use $S_E(b)$ for evidence support, $D(b,C)$ for evidence drift, and $L(a,C')$ for loss of truth-critical content.

\paragraph{Invalid-rewrite diagnostics.}
For rewrites that fail the binary gate, AtomEval reports:
\begin{equation}
\operatorname{EvDrift}
=\mathbb{I}\!\left[
\mathcal{H}=0
\land
\exists b\in A': S_E(b)\land D(b,C)
\right].
\end{equation}

\begin{equation}
\operatorname{ScopeLoss}
=\mathbb{I}\!\left[
\mathcal{H}=0
\land
\exists a\in A: L(a,C')
\right].
\end{equation}
Here, $D(b,C)$ means that $b$ overlaps with the original claim target but replaces rather than preserves the attacked proposition.
$L(a,C')$ means that the rewrite deletes, weakens, or generalizes truth-critical content from $a$, such as dates, quantities, negation, modality, exclusivity, named entities, or relation-specific constraints.

\paragraph{Valid-rewrite diagnostics.}
For rewrites that pass the binary gate, AtomEval reports:
\begin{equation}
\begin{aligned}
\operatorname{EvEnt}
&=\mathbb{I}\!\left[
\mathcal{H}=1
\land
\exists b\in \Delta: S_E(b)
\right],\\
\operatorname{UnverAdd}
&=\mathbb{I}\!\left[
\mathcal{H}=1
\land
\exists b\in \Delta: \neg S_E(b)
\right].
\end{aligned}
\end{equation}
EvEnt captures evidence-supported contextual additions.
UnverAdd captures unsupported factual additions.
Both are compatible with AtomEval's core validity gate because the gate evaluates attacked-proposition preservation rather than full sentence-level semantic equivalence. We therefore report them as separate diagnostics rather than treating them as automatic validity failures.

\paragraph{Reporting.}
Diagnostics are binary and non-exclusive.
EvDrift and ScopeLoss are reported over invalid rewrites, while EvEnt and UnverAdd are reported over valid rewrites.
Implementation details for the NLI verifier, evidence support, and constraint checks are provided in Appendix~\ref{app:Implement}.

\section{Implementation Details}
\label{app:Implement}

All experiments were implemented using PyTorch and the Hugging Face Transformers library~\cite{wolf19huggingface}.

\paragraph{Victim verifiers.}
We evaluate Gemma-7B-based~\cite{gemmateam2024gemma} and BERT-base-uncased~\cite{devlin2019bert} fact verification models, both fine-tuned on FEVER claim--evidence pairs.

\paragraph{AtomEval verifier.}
AtomEval uses a frozen DeBERTa-v3 NLI model~\cite{He21DeBERTaV3} for the binary validity gate and evidence-support diagnostics.
The checkpoint is \texttt{MoritzLaurer/DeBERTa-v3-base-mnli-fever}\\
\texttt{-anli}.
The gate checks whether rendered original atoms are entailed by the adversarial rewrite, with additional consistency checks over dates, quantities, months, and exclusivity markers.

\paragraph{Surface metrics.}
SBERT similarity is computed with \texttt{all-mpnet-base-v2}~\cite{Reimers19Sentence}, and PPL is measured using GPT-2~\cite{radford19language}.
These metrics are used only as comparison proxies.

\paragraph{Hardware.}
Experiments were conducted on a single NVIDIA A100-SXM4 GPU with 80GB memory.

\section{Attack Prompt Templates}
\label{app:prompts}

Adversarial rewrites are generated using a zero-shot prompting setup.
Each prompt contains a shared system instruction, a strategy-specific rewriting instruction, and the input claim--evidence pair.
The evidence is provided to the generator but is kept fixed during verification.
All strategies require the generator to output exactly one rewritten claim without explanations or labels.

Figure~\ref{fig:prompt_template} shows the shared prompt structure and one representative strategy template for Add.Unver./AdvAdd-style unsupported addition.
Other attack families use the same system instruction with strategy-specific guidance.

\begin{figure*}[t]
\centering
\small
\begin{tcolorbox}[
    colback=white,
    colframe=black!80,
    arc=1.2mm,
    title=\textbf{Representative Zero-shot Attack Prompt},
    fonttitle=\sffamily\bfseries,
    left=3mm, right=3mm, top=2mm, bottom=2mm
]

\noindent\textbf{System instruction} \\
\texttt{You are an expert in adversarial evaluation for automated fact verification.}

\medskip
\noindent
\texttt{You will be given an Original Claim, and fixed Evidence.} \\
\texttt{Your task is to rewrite the claim according to the specified attack strategy.}

\medskip
\noindent
\texttt{General rules:}
\begin{itemize}[leftmargin=1.4em, noitemsep, topsep=2pt]
    \item \texttt{Output exactly ONE rewritten claim.}
    \item \texttt{Output a single sentence whenever possible.}
    \item \texttt{Do not output explanations, labels, or prefixes.}
    \item \texttt{Keep the claim fluent, natural, and fact-checkable.}
    \item \texttt{Avoid copying the original claim or an evidence sentence verbatim.}
    \item \texttt{The evidence is fixed and must not be rewritten.}
\end{itemize}

\medskip
\noindent\textbf{Input} \\
\texttt{Original Claim: \{original\_claim\}} \\
\texttt{Evidence: \{evidence\}}

\medskip
\noindent\textbf{Strategy instruction: Add.Unver./AdvAdd-style unsupported addition} \\
\texttt{Rewrite the original claim by preserving its main assertion and adding exactly one plausible auxiliary detail.}
\texttt{The added detail should be related to the original claim, but it should not be directly stated or explicitly entailed by the fixed evidence.}
\texttt{It should sound like a normal background detail that would require additional verification beyond the provided evidence.}

\medskip
\noindent
\texttt{Important constraints:}
\begin{itemize}[leftmargin=1.4em, noitemsep, topsep=2pt]
    \item \texttt{Preserve the original claim's main assertion.}
    \item \texttt{Add exactly one minor auxiliary detail.}
    \item \texttt{Do not copy the added detail from the evidence.}
    \item \texttt{Do not recombine multiple evidence facts; that belongs to fact mixing.}
    \item \texttt{Do not correct the original claim.}
    \item \texttt{Keep the rewritten claim as one fluent declarative sentence.}
\end{itemize}

\medskip
\noindent
\texttt{Rewritten claim:}

\end{tcolorbox}
\vspace{-1mm}
\caption{
Representative zero-shot prompt used for adversarial claim rewriting.
The shared system instruction is combined with a strategy-specific rewriting instruction and the fixed claim--evidence input.
}
\label{fig:prompt_template}
\end{figure*}

\section{Diagnostic-Guided Repair Probe}
\label{app:repair_prompt}

\paragraph{Repair prompt.}
For each invalid raw-successful rewrite, we use the following prompt:

\begin{tcolorbox}[
    colback=white,
    colframe=black!60,
    arc=0.8mm,
    boxrule=0.4pt,
    left=1.5mm, right=1.5mm, top=1mm, bottom=1mm,
    title=\textbf{Diagnostic-Guided Repair Instruction},
    fonttitle=\footnotesize\sffamily\bfseries
]
\footnotesize
Edit the invalid rewrite directly, using the original claim as the semantic anchor and the AtomEval diagnosis as repair guidance.
Minimally fix the diagnosed failure while preserving the rewrite's wording and structure.
Restore any changed or missing truth-critical constraint, including entities, relations, numbers, dates, negation, exclusivity terms, roles, locations, and modifiers.
Do not correct toward the evidence, replace the attacked proposition with evidence-supported facts, or add caveats or truth-status labels.
Return one declarative sentence as JSON:
\texttt{\{"repair\_claim": "..."\}}.
\end{tcolorbox}

\section{Additional Results and Case Studies}
\label{app:additional_results}

\subsection{Full Surface Results}
\label{app:surface_metrics}

Table~\ref{tab:bert_surface_metrics} reports surface-proxy values for successful attacks on the BERT-based verifier. GPT-4.1 generally yields higher SBERT similarity than Mistral-7B, while Adv.PPL varies substantially by attack family. 

\begin{table}[t]
\centering
\small
\setlength{\tabcolsep}{5pt}
\begin{tabular}{llcc}
\toprule
\textbf{Attack} & \textbf{Gen.} & \textbf{SBERT}$_\uparrow$ & \textbf{Adv.PPL}$_\downarrow$ \\
\midrule
AdvAdd & GPT-4.1     & 0.903 & 65.59 \\
AdvAdd & Mistral-7B & 0.723 & 47.35 \\
\midrule
Colloquial & GPT-4.1     & 0.880 & 128.97 \\
Colloquial & Mistral-7B & 0.734 & 132.16 \\
\midrule
DeSePtion & GPT-4.1     & 0.803 & 38.61 \\
DeSePtion & Mistral-7B & 0.686 & 35.29 \\
\midrule
FactMix & GPT-4.1     & 0.753 & 78.96 \\
FactMix & Mistral-7B & 0.657 & 64.22 \\
\midrule
Omission & GPT-4.1     & 0.840 & 186.46 \\
Omission & Mistral-7B & 0.571 & 150.23 \\
\bottomrule
\end{tabular}
\caption{
Surface-proxy statistics for successful adversarial rewrites on the BERT-based verifier.
SBERT reports the mean Sentence-BERT similarity between the original claim and the successful rewrite.
Adv.PPL reports the mean GPT-2 perplexity of successful adversarial rewrites.
}
\label{tab:bert_surface_metrics}
\end{table}

\subsection{Full Diagnostic Result}
\label{app:full_diagnostic}

Table~\ref{tab:full_atomeval_breakdown} shows that invalid successful attacks are dominated by ScopeLoss, especially for omission-style rewrites, while evidence drift is most prominent in FactMix and colloquial generations.

\begin{table*}[t]
\centering
\scriptsize
\setlength{\tabcolsep}{2.6pt}
\renewcommand{\arraystretch}{1.05}
\resizebox{\linewidth}{!}{
\begin{tabular}{ll l r r r r r r r}
\toprule
\textbf{Verifier} & \textbf{Generator} & \textbf{Attack Family} &
\textbf{N} &
\textbf{Invalid} &
\textbf{Valid} &
\textbf{EvDrift} &
\textbf{ScopeLoss} &
\textbf{EvEnt} &
\textbf{UnverAdd} \\
\midrule

BERT-based & GPT-4.1 & \textsc{Add.Unver./AdvAdd} & 37 & 0.0 (0) & 100.0 (37) & -- & -- & 0.0 (0) & 97.3 (36) \\
BERT-based & GPT-4.1 & \textsc{Colloquial/Lexical} & 42 & 19.0 (8) & 81.0 (34) & 50.0 (4) & 100.0 (8) & 2.9 (1) & 20.6 (7) \\
BERT-based & GPT-4.1 & \textsc{DeSePtion-style} & 189 & 9.5 (18) & 90.5 (171) & 22.2 (4) & 94.4 (17) & 32.8 (56) & 71.9 (123) \\
BERT-based & GPT-4.1 & \textsc{GEM-style FactMix} & 103 & 57.3 (59) & 42.7 (44) & 91.5 (54) & 89.8 (53) & 86.4 (38) & 79.5 (35) \\
BERT-based & GPT-4.1 & \textsc{Omission-style} & 83 & 94.0 (78) & 6.0 (5) & 16.7 (13) & 98.7 (77) & 0.0 (0) & 20.0 (1) \\

\midrule

BERT-based & Mistral-7B & \textsc{Add.Unver./AdvAdd} & 129 & 67.4 (87) & 32.6 (42) & 48.3 (42) & 95.4 (83) & 30.9 (13) & 100.0 (42) \\
BERT-based & Mistral-7B & \textsc{Colloquial/Lexical} & 85 & 62.4 (53) & 37.6 (32) & 73.6 (39) & 86.8 (46) & 25.0 (8) & 18.8 (6) \\
BERT-based & Mistral-7B & \textsc{DeSePtion-style} & 298 & 83.2 (248) & 16.8 (50) & 63.3 (157) & 94.0 (233) & 72.0 (36) & 78.0 (39) \\
BERT-based & Mistral-7B & \textsc{GEM-style FactMix} & 194 & 75.8 (147) & 24.2 (47) & 81.6 (120) & 85.0 (125) & 85.1 (40) & 83.0 (39) \\
BERT-based & Mistral-7B & \textsc{Omission-style} & 165 & 100.0 (165) & 0.0 (0) & 41.8 (69) & 99.4 (164) & -- & -- \\

\midrule

Gemma-7B & GPT-4.1 & \textsc{Add.Unver./AdvAdd} & 11 & 0.0 (0) & 100.0 (11) & -- & -- & 0.0 (0) & 100.0 (11) \\
Gemma-7B & GPT-4.1 & \textsc{Colloquial/Lexical} & 64 & 85.9 (55) & 14.1 (9) & 81.8 (45) & 67.3 (37) & 11.1 (1) & 22.2 (2) \\
Gemma-7B & GPT-4.1 & \textsc{DeSePtion-style} & 30 & 6.7 (2) & 93.3 (28) & 50.0 (1) & 100.0 (2) & 35.7 (10) & 60.7 (17) \\
Gemma-7B & GPT-4.1 & \textsc{GEM-style FactMix} & 62 & 50.0 (31) & 50.0 (31) & 100.0 (31) & 83.9 (26) & 96.8 (30) & 67.7 (21) \\
Gemma-7B & GPT-4.1 & \textsc{Omission-style} & 154 & 94.8 (146) & 5.2 (8) & 24.0 (35) & 99.3 (145) & 0.0 (0) & 12.5 (1) \\

\midrule

Gemma-7B & Mistral-7B & \textsc{Add.Unver./AdvAdd} & 157 & 84.1 (132) & 15.9 (25) & 55.3 (73) & 89.4 (118) & 48.0 (12) & 96.0 (24) \\
Gemma-7B & Mistral-7B & \textsc{Colloquial/Lexical} & 152 & 88.2 (134) & 11.8 (18) & 85.1 (114) & 85.8 (115) & 66.7 (12) & 38.9 (7) \\
Gemma-7B & Mistral-7B & \textsc{DeSePtion-style} & 256 & 87.1 (223) & 12.9 (33) & 70.9 (158) & 92.8 (207) & 78.8 (26) & 78.8 (26) \\
Gemma-7B & Mistral-7B & \textsc{GEM-style FactMix} & 139 & 88.5 (123) & 11.5 (16) & 91.1 (112) & 86.2 (106) & 87.5 (14) & 62.5 (10) \\
Gemma-7B & Mistral-7B & \textsc{Omission-style} & 294 & 99.7 (293) & 0.3 (1) & 45.4 (133) & 99.0 (290) & -- & -- \\

\bottomrule
\end{tabular}
}
\caption{
AtomEval diagnostics for raw successful attacks.
Percentages are shown with counts in parentheses.
Invalid and Valid denote rewrites that fail or pass the binary gate; valid-side diagnostics are omitted for fewer than five valid rewrites.
}
\label{tab:full_atomeval_breakdown}
\end{table*}

\subsection{Case Study}
\label{app:case_study}

We provide a claim-centered case study illustrating how different attack strategies and generators can produce qualitatively different rewrites for the same original instance.

\noindent\textbf{Original claim:} ``Danger UXB is a desk.''

\noindent\textbf{Evidence:} ``Danger UXB is a 1979 British ITV television series set during the Second World War developed by John Hawkesworth and starring Anthony Andrews as Lieutenant Brian Ash, an officer in the Royal Engineers.''

\begin{table*}[t]
\centering
\scriptsize
\setlength{\tabcolsep}{3pt}
\renewcommand{\arraystretch}{1.12}
\resizebox{\linewidth}{!}{
\begin{tabular}{p{0.09\linewidth} p{0.07\linewidth} p{0.36\linewidth} p{0.07\linewidth} p{0.12\linewidth} p{0.25\linewidth}}
\toprule
\textbf{Strategy} &
\textbf{Gen.} &
\textbf{Adversarial Rewrite} &
\textbf{Attack} &
\textbf{Observation} &
\textbf{Explanation} \\
\midrule

\multirow{2}{*}{\textsc{Lexical}}
& GPT-4.1 &
``Danger UXB is a piece of furniture.'' &
Failure &
ScopeLoss &
The rewrite generalizes ``desk'' to ``piece of furniture.'' It weakens the original proposition and does not flip the verifier in this instance. \\

& Mistral &
``Danger UXB's actually a telly show, mate. It's a 1979 British series, see? Set during WW2, and starring Tony Andrews as an officer in the Royal Engineers.'' &
Success &
EvDrift &
The rewrite replaces the original false desk proposition with the evidence-supported fact that \textit{Danger UXB} is a television series. \\

\midrule

\multirow{2}{*}{\textsc{DeSePtion}}
& GPT-4.1 &
``Danger UXB, which shares its name with a well-known British television series, refers to a type of desk introduced in the late 20th century.'' &
Success &
Valid + EvEnt + UnverAdd &
The rewrite preserves the original false desk proposition while adding evidence-supported context about the television series. \\

& Mistral &
``During the Second World War, there existed a British television series titled Danger UXB, developed by John Hawkesworth and starring Anthony Andrews as Lieutenant Brian Ash, whose duties were not limited solely to a desk.'' &
Success &
EvDrift + ScopeLoss &
The rewrite shifts toward the evidence-supported television-series description and weakens the desk proposition into a peripheral phrase. \\

\midrule

\multirow{2}{*}{\textsc{FactMix}}
& GPT-4.1 &
``Danger UXB is a desk used by the Royal Engineers during the Second World War.'' &
Failure &
Valid + UnverAdd &
The rewrite preserves the false desk proposition but adds unsupported details blended from evidence-related entities and setting information. It does not flip the verifier here. \\

& Mistral &
``Anthony Andrews, known for his role in the 1979 ITV series Danger UXB, appeared as a desk officer in the Royal Engineers during the Second World War.'' &
Success &
EvDrift + ScopeLoss &
The rewrite shifts the subject from \textit{Danger UXB} to Anthony Andrews, drops the original desk proposition, and mixes evidence-supported details with an unsupported desk-related role. \\

\midrule

\multirow{2}{*}{\textsc{Omission}}
& GPT-4.1 &
``Danger UXB is a television series.'' &
Success &
EvDrift &
The rewrite directly replaces the original false desk proposition with the evidence-supported television-series fact. \\

& Mistral &
``A device used for detecting and dealing with explosives is often a TV series setting.'' &
Success &
ScopeLoss &
The rewrite generalizes away from the named entity and the original desk proposition, so the attacked proposition is no longer recoverable. \\

\midrule

\multirow{2}{*}{\textsc{AdvAdd}}
& GPT-4.1 &
``Danger UXB is a desk that was once used in a British government office.'' &
Failure &
Valid + EvEnt + UnverAdd&
The rewrite preserves the original false desk proposition while adding an unsupported office-use detail. \\

& Mistral &
``Danger UXB was frequently used on the battlefields of Europe during the Second World War.'' &
Failure &
ScopeLoss &
The rewrite drops the original desk proposition and introduces an unsupported battlefield-use interpretation of \textit{Danger UXB}; it is semantically altered but does not flip the verifier. \\

\bottomrule
\end{tabular}
}
\caption{
Claim-centered case study under fixed evidence.
Strategy abbreviations correspond to the five attack families in Section~\ref{sec:attack_families}: Lexical, DeSePtion, FactMix, Omission, and AdvAdd.
The examples show that attack success and proposition preservation are distinct: some successful attacks are invalid because they drift toward evidence or lose the original proposition, while valid rewrites may still introduce evidence-supported or unsupported additional content.
Diagnostic tags such as EvEnt and UnverAdd characterize these expansions but do not affect VASR.
}
\label{tab:case_study_appendix}
\end{table*}